\pdfoutput=1
\documentclass[11pt]{article}

\usepackage{EMNLP2023}
\usepackage{times}
\usepackage{latexsym}

\usepackage{graphicx}
\usepackage{amsmath}
\usepackage{amsfonts}
\usepackage{amssymb}
\usepackage{hyperref} 
\usepackage{booktabs} 
\usepackage{multirow}
\usepackage{colortbl}
\usepackage{makecell}
\usepackage{float}
\usepackage[T1]{fontenc}
\usepackage[utf8]{inputenc}

\usepackage{microtype}

\usepackage{inconsolata}

\title{PromptNER: A Prompting Method for Few-shot Named Entity Recognition via $k$ Nearest Neighbor Search}

\author{Mozhi Zhang, Hang Yan, Yaqian Zhou, Xipeng Qiu \\
 School of Computer Science, Fudan University\\
 Shanghai Key Laboratory of Intelligent Information Processing, Fudan University \\
  \texttt{mzzhang22@m.fudan.edu.cn} \\
  \texttt{{hyan19, zhouyaqian, xpqiu}@fudan.edu.cn} \\}

\begin{document}
\maketitle
\begin{abstract}
Few-shot Named Entity Recognition (NER) is a task aiming to identify named entities via limited annotated samples. 
Recently, prototypical networks have shown promising performance in few-shot NER. 
Most of prototypical networks will utilize the entities from the support set to construct label prototypes and use the query set to compute span-level similarities and optimize these label prototype representations. 
However, these methods are usually unsuitable for fine-tuning in the target domain, where only the support set is available.
In this paper, we propose PromptNER: a novel prompting method for few-shot NER via $k$ nearest neighbor search. 
We use prompts that contains entity category information to construct label prototypes, which enables our model to fine-tune with only the support set.
Our approach achieves excellent transfer learning ability, and extensive experiments on the Few-NERD and CrossNER datasets demonstrate that our model achieves superior performance over state-of-the-art methods.
\end{abstract}

\section{Introduction}
% 为什么有few-shot ner
Named Entity Recognition (NER) is a fundamental NLP task to extract entities from unstructured text. In traditional fully supervised NER scenarios, deep neural architectures~\cite{huang2015bidirectional, lample2016neural, ma2016end, yan2019tener} have shown great ability to recognize entities with sufficient human-annotated data. However, acquiring such human-annotated data can be expensive and time-consuming since the demand for domain-specific knowledge. Previous NER models usually struggle to leverage very limited labeled data to recognize entities in practical scenarios owing to these data-hungry characteristics. Furthermore, the classifier head of a traditional NER system needs to be retrained from scratch when the number or type of entity class changes.
\begin{figure}[!ht]
  \centering
  \includegraphics[width=0.95\columnwidth]{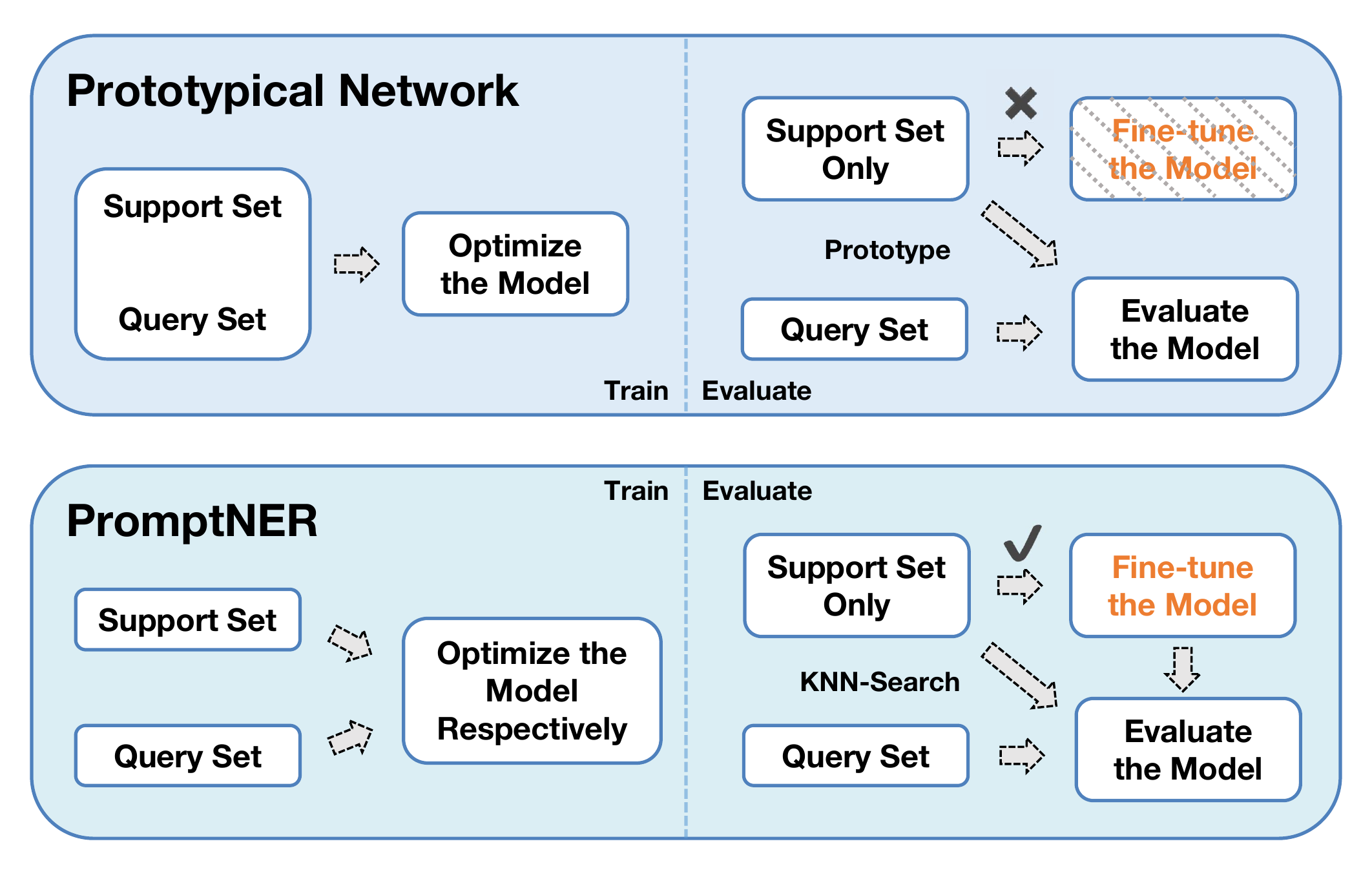}
  \vspace{-0.3em}
  \caption{The difference between traditional Prototypical Networks and PromptNER.}  \label{fig:difference}
\end{figure}
Therefore, few-shot NER has drawn much attention in the information extraction field. 

% few-shot ner早期工作 token-level的model，以及label semantic NER
Owing to only a few labeled examples (usually called support examples) available,~\citet{fritzler2019few} and~\citet{hou-etal-2020-shot} propose to compute token-level similarities between the label prototypes or each token of support sets and each token of query sets. Based on previous works,~\citet{das-etal-2022-container} propose CONTaiNER, the first method using contrastive learning to enhance the token representation of PTMs for few-shot NER.~\citet{ma-etal-2022-label} propose an architecture consisting of two pre-trained encoders to encode the sentence and label words, proving effective for low-resource NER.

% 介绍span-based的few-shot ner
Recently, span-based NER~\cite{yu2020named, li-etal-2020-unified, yan2022embarrassingly} has demonstrated exemplary performance in various NER tasks.~\citet{ma-etal-2022-decomposed} decomposes the few-shot NER task into two distinct stages, i.e. span-detection and entity-typing. They also use MAML~\cite{finn2017model}, a meta-learning algorithm, to enhance the performance of their model.~\citet{wang-etal-2022-enhanced} converts the NER task into a span-matching problem and propose a novel span refining module which applies the Soft-NMS~\cite{bodla2017soft, shen-etal-2021-locate} algorithm during beam search. These span-based prototypical networks achieve significant improvements over token-level few-shot NER baselines, which avoid the token-level label dependency problem.

% span-based prototypical network的缺点
Despite the promising performance of span-based prototypical networks. Two problems limit these methods.
1)~The span-level metric learning of the prototypical network is based on support sets and query sets, where samples from support sets are used to construct the label prototypes, and query samples are used to compute the span-level similarities and optimize these label prototypes.
However, only the label of support samples is available in the test scenario. Previous prototypical networks~\cite{fritzler2019few, wang-etal-2022-enhanced, wang-etal-2022-spanproto} usually do not update any parameter of their models on the novel support set, which limits the transfer learning capability of these methods. 
2)~Previous span detectors usually extract some false positive spans.
In few-shot NER, unseen new classes in the test set are usually tagged as O-type~\cite{das-etal-2022-container} during training. 
Unfortunately, previous span-based models are class-agnostic in the span-detection stage. It is challenging to detect unseen new class span only during the span detection stage since these models have been thoroughly trained in the source domain to regard the unseen new class entities as O-type. 
To address this problem, ~\citet{ma-etal-2022-decomposed} and \citet{wang-etal-2022-spanproto} filter some false positive spans, which are too far from label prototypes.~\citet{wang-etal-2022-enhanced} introduce an O-type prototype to match false positive spans in the query set. 
However, owing to the limited support examples, label prototypes constructed by support samples may not precisely represent the class distribution in the feature space. 

% 针对上述问题，我们改进的方法
This paper proposes PromptNER: a simple but effective prompting method for few-shot NER. 
First, we construct a natural language prompt to instruct Pre-trained Language Models (PLMs) to extract entities with specific classes.
Then we design a position-aware biaffine module for recalling candidate spans and a prompt-based classifier for entity typing.
Inspired by~\citet{wang2022k}, we introduce $k$ nearest neighbor search to leverage the ground truth entity representations from support examples. 
\begin{figure}[!ht]
  \centering
  \includegraphics[width=0.95\columnwidth]{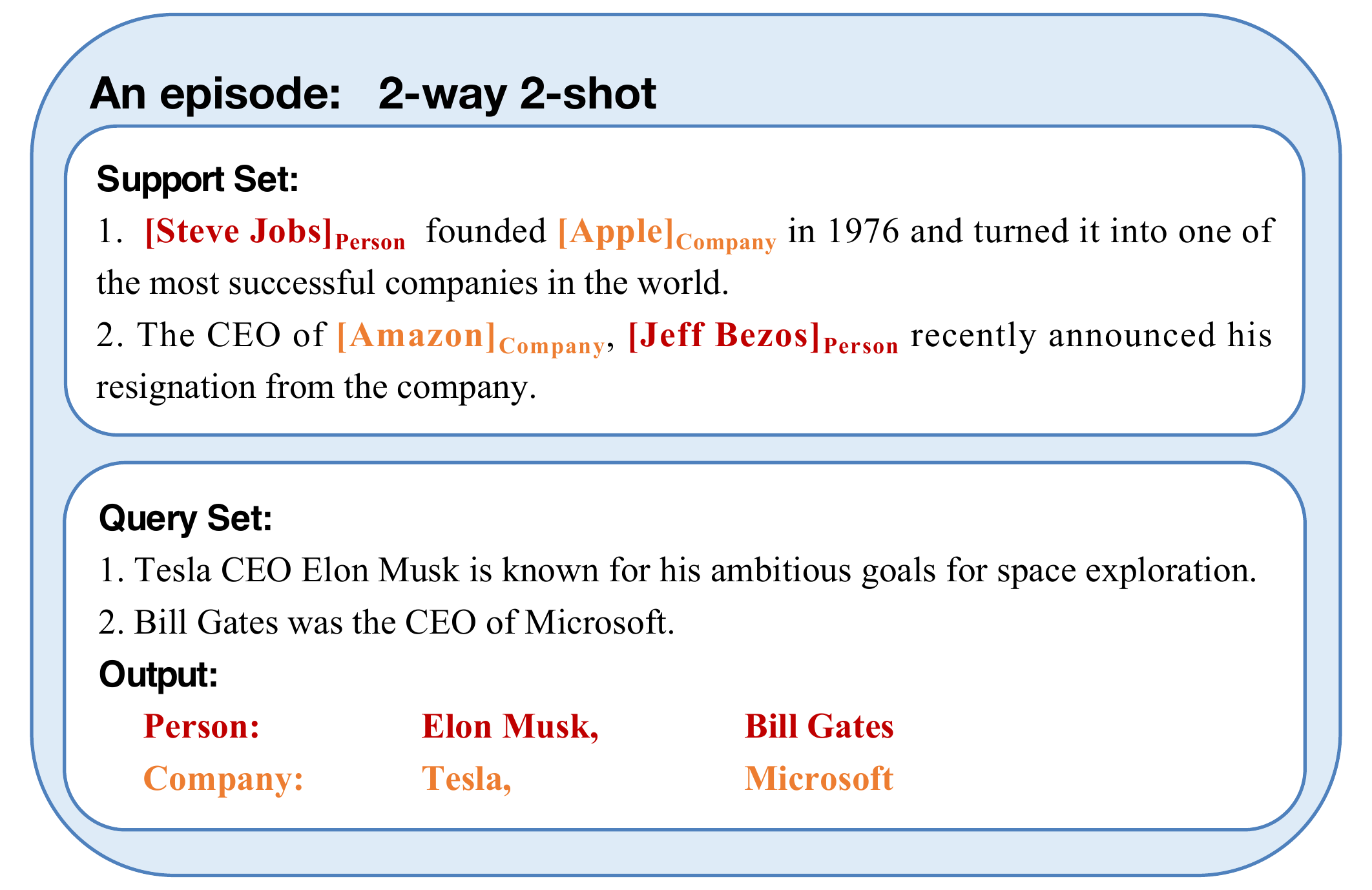} 
  \vspace{-0.3em}
  \caption{An example of 2-way 2-shot episode.} \label{fig:episdoe}
\end{figure}
The difference between typical prototypical networks and our method is shown in Figure~\ref{fig:difference}. Unlike previous prototypical networks, the optimization process of our model is not limited to the support set and query set format. Like traditional NER, our model only requires sentences and their corresponding label sentences for training. Therefore, we can fine-tune our model on a novel support set without gaps between the training and fine-tuning stages.
We alse propose a novel rerank strategy to filter false positive spans, 
% 效果如何
We evaluate PromptNER on multiple benchmark datasets, including Few-NERD~\cite{ding-etal-2021-nerd} and CrossNER~\cite{hou-etal-2020-shot}. The experimental results demonstrate that PromptNER achieves superior performance over state-of-the-art few-shot NER methods and the effectiveness of the rerank strategy and fine-tuning stage.

\section{Problem Formulation}
In this part, we formally introduce the problem formulation of few-shot named entity recognition~(NER).

% 输入输出和supervised 是一样的
Similarly to the supervised NER system, the input of the few-shot NER system is a natural language sentence $X$ which contains $n$ words. And the output $Y = \{y_{i}\}_{i=1}^{n} $ is a label sentence, where $y_i \in \mathcal{T}$, $\mathcal{T}$ is the entity type set with O-type~(Outside).
% N-way K-shot
Following ~\citet{ding-etal-2021-nerd}, we adapt the standard N-way K-shot setting to train and evaluate the few-shot NER system. During training, each episode data $\varepsilon_{train} = \{\mathcal{S}_{train}, \mathcal{Q}_{train}, \mathcal{T}_{train}\}$ contains a support set $\mathcal{S}_{train}$, a query set $\mathcal{Q}_{train}$ and entity type set $\mathcal{T}_{train}$. A support or query set contains $N$ classes ($N$-way) and $K$ examples ($K$-shot) for each entity class respectively, where  $\mathcal{S}_{train} \cap \mathcal{Q}_{train} = \varnothing$. For testing, we utilize a novel episode  $\varepsilon_{test} = \{\mathcal{S}_{test}, \mathcal{Q}_{test}, \mathcal{T}_{test}\}$ to evaluate the few-shot NER system, where $\mathcal{T}_{train} \cap \mathcal{T}_{test} =$~O.A typical 2-way 2-shot episode is shown in Figure~\ref{fig:episdoe}.

\begin{figure*}[t]
  \centering
  \includegraphics[width=\textwidth]{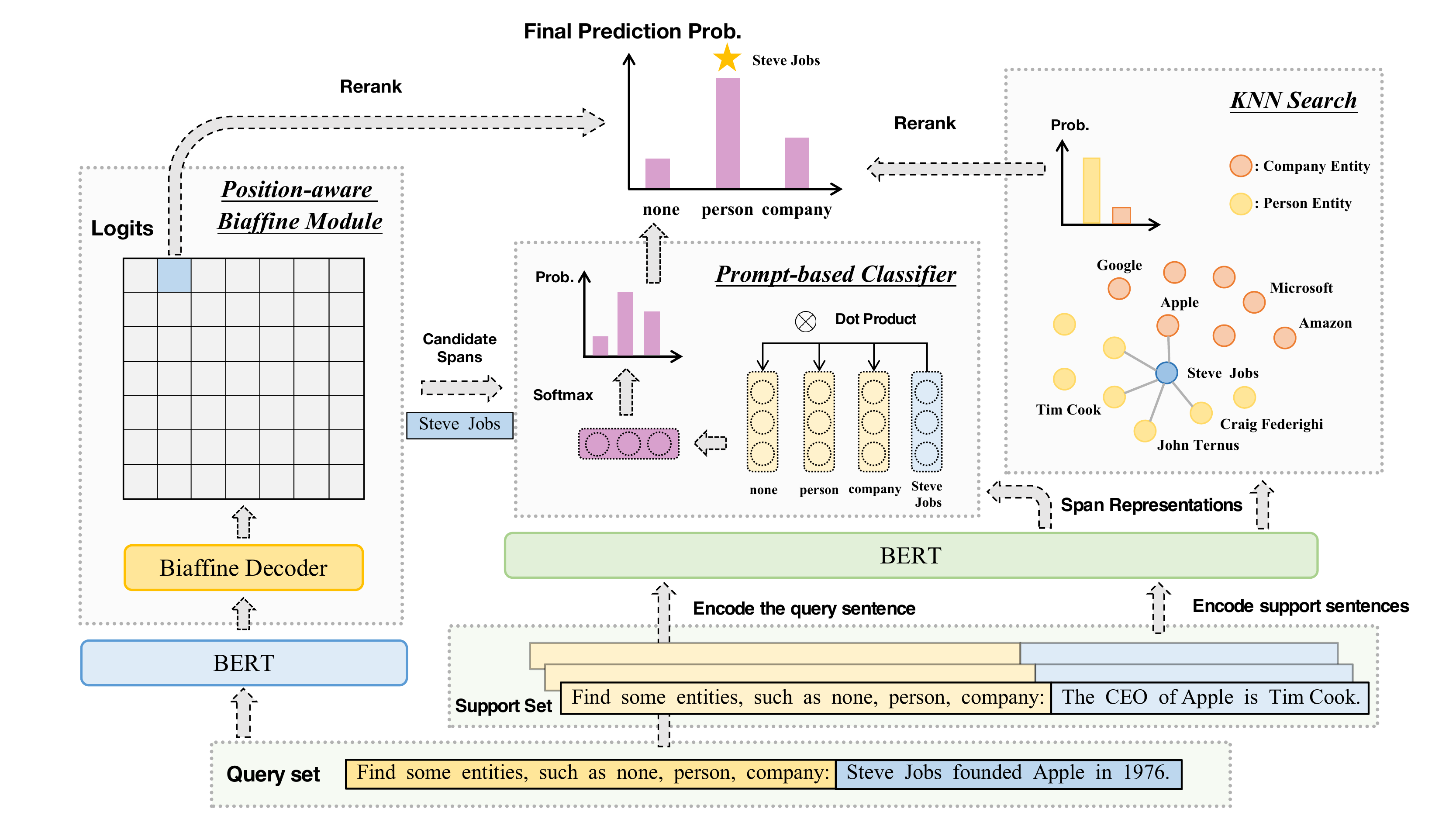}
  \caption{Model structure of our method.} \label{fig:strucutre}
\end{figure*}
\section{Proposed Method}
In this section, we formally present our proposed PromptNER. The architecture of PromptNER is shown in the Figure~\ref{fig:strucutre}. 
\subsection{Input Construction}
% 输入是什么，type set怎么定义
Formally, the input of a NER system is a natural language sentence. Given a sentence consisting of $n$ words $X = \left[x_1, x_2, ..., x_n \right]$ and an entity type set $\mathcal{T} = \{none, t_1, t_2, ..., t_{m-1}\}$ where $m = \left\vert \mathcal{T} \right\vert$ and $none$ means O-type, we use the pre-defined prompt template to reconstruct the input sentence as follows:
\begin{align}
  X_p &= \mathrm{F}_{prompt}(\mathcal{T})\oplus X, \nonumber \\
       &= \left[X_l, X_m, X\right] \label{ali: one}
\end{align}
% 举个例子
where $\mathrm{F}_{prompt}(\mathcal{T})$ is a function which fills the template using the entity type set $\mathcal{T}$. 
For example, suppose the entity type set is $ \{none, person, company\}$, and the input sentence is ``Steve Jobs founded Apple in 1976.''. The reconstructed input, using the template ``Find some entities, such as $none, t_1, t_2, ..., t_{m-1}$: '' will be ``Find some entities, such as none, person, company: Steve Jobs founded Apple in 1976.''. Additionally, the input $X_p$ could also be split into three parts shown in (\ref{ali: one}), where $X_l = $ ``Find some entities, such as'', $X_m = $ ``none, person, company''.\footnote{The reason why we split $X_p$ into three parts in (\ref{ali: one}) is that we only use the embedding of label words and words from the original input sentence. And $l, m, n$ are the word number of each part.}~The reconstructed input $X_p$ provides label information to the model and instructs the model to extract some entities mentioned in the prompt.

% 这里介绍怎么判断是否为一个span
\subsection{Position-aware Biaffine Module}\label{sec:two_spec}
We follow~\citet{yu2020named} and ~\citet{yan2022embarrassingly} to convert the span-detection task into a binary classification task. For a sentence with $n$ tokens, we need to perform binary classification task $n(n+1)/2$ times. To this end, our method first uses a pre-trained encoder to encode the prompt and the input sentence:

\begin{align}
  \mathbf{H} &= \mathrm{Encoder}(X_p), \nonumber \\
             &=  \left[\mathbf{H}_l, \mathbf{H}_m, \mathbf{H}_n\right]  \label{ali: two}
\end{align}
where $\mathbf{H} \in \mathcal{R}^{(l + m + n) \times d}$, and $d$ is the embedding size. The encoder is typically a pre-trained language model, such as BERT~\cite{devlin-etal-2019-bert}, RoBERTa~\cite{liu2019roberta}. Because several words may be tokenized into some subwords, we use mean-pooling to obtain the representation of each word. 
Meanwhile, $\mathbf{H}_l$ will be ignored, and the label word embedding $\mathbf{H}_m$ is utilized in the prompt-based classifier, which will be illustrated in Section~\ref{sec:three_spec}.

Then we design a biaffine model which incorporates absolute and relative position information. Inspired by~\citet{su2022global}, we apply RoPE into the span detection stage to inject absolute and relative position information, which satisfies the constraint $\mathcal{R}^\top_i\mathcal{R}_j = \mathcal{R}_{j-i}$. For a span that ranges from $i$-th word to $j$-th word, we can calculate the prediction logit as follows: 
\begin{align}
  \mathbf{h}_s & = \mathrm{LeakyReLU}(\mathbf{h}_iW_s), \nonumber \\
  \mathbf{h}_e & = \mathrm{LeakyReLU}(\mathbf{h}_jW_e),  \nonumber \\
  \mathbf{R}_{i,j} & = \mathbf{h}_s^\top  \mathrm{U} \mathbf{h}_e+  (\mathcal{R}_{i}\mathbf{h}_iW_p)^\top(\mathcal{R}_{j}\mathbf{h}_jW_p), \nonumber \\
  & = \mathbf{h}_s^\top\mathrm{U}\mathbf{h}_e + (\mathbf{h}_iW_p)^\top\mathcal{R}_{j-i}(\mathbf{h}_jW_p)  
\end{align}
where  $W_s, W_e, W_p\in \mathcal{R}^{d \times h}$, $\mathrm{U}, \mathcal{R}_{i}, \mathcal{R}_{j} \in \mathcal{R}^{h \times h}$, and $h$ is the hidden size. For a sentence with $n$ words, we can get a score matrix $\mathbf{R} \in \mathcal{R}^{n \times n}$. We mask the lower triangle part of $\mathbf{R}$ (where $i > j$), to filter all the impossible spans which contain words from $i$-th to $j$-th. To address the issue of sample imbalance, we use the span-based class imbalance loss proposed in~\citet{su2022global}:
\begin{align}
    \mathcal{L}_{pos} &= \log \Big( 1 + \sum_{\left(i, j\right) \in \mathcal{S}_{pos}} e^{-r\left(i, j \right)}\Big), \nonumber\\  
    \mathcal{L}_{neg}  &= \log \Big( 1 + \sum_{\left(i, j\right) \in \mathcal{S}_{neg}} e^{r\left(i, j \right)}\Big),  \nonumber \\
    \mathcal{L}_{span} &= \mathcal{L}_{pos} + \mathcal{L}_{neg}, \nonumber
\end{align}
where $1 \leq i \leq j\leq n$, $\mathcal{S}_{pos} = \left\{\left(s_k, e_k\right)\right\}_{k=1}^{N}$ represents the collection of candidates spans(noun phrase), and $N$ is the entity span number of the input sentence. $\mathcal{S}_{neg}$ represents the collection of spans which not belong to noun phrases accordingly. 

During inference, we extract with the top-$3k$ logits from the upper triangle part of score matrix $\mathbf{R}$ to recall more candidate spans, where $k$ corresponds to the $k$-shot setting. 

% 这里介绍怎么对span进行分类
\subsection{Prompt-based Classifier}\label{sec:three_spec}
In this section, we propose a novel approach to classify each candidate span.% extracted from the position-aware biaffine module.
~Unlike the technique presented by~\citet{ma-etal-2022-label}, our method incorporates the semantic information of the input sentence into the label embedding. Moreover, we introduce an additional embedding type for the ``none'' category, which assists in identifying and filtering out some false positive spans.

\subsubsection{Classification with Prompt}
For each example $\left( X, Y, \mathcal{T}\right)$ in $\mathcal{D}_{train}$, we utilize $\mathbf{H}_m, \mathbf{H}_n$ computed in (\ref{ali: two}) to compute the classification probability of each entity span in $\mathcal{S}_{pos}$. Specifically, for the i-th span $\left(s_i, e_i\right)$ in $\mathcal{S}_{pos}$, we can obtain its representation as follows:
\begin{align}
    \mathbf{u_i} = \frac{1}{e_i-s_i+1}\sum_{k=s_i}^{e_i} \mathbf{h}_k, \nonumber 
\end{align}
where $\mathbf{h}_k \in \mathbf{H}_n$, $s_i, e_i$ denote the starting and ending indices for the i-th span, respectively.

The probability distribution can be calculated as follows:
\begin{align}
  p(y|s_i, e_i) =  \mathrm{Softmax}(\frac{ \mathbf{H}_m \mathbf{u}_i^\top}{\sqrt{d}}), \nonumber
\end{align}
where $\mathbf{H}_m \in \mathcal{R}^{m \times d}$ , $m$ is the class number and $d$ is the embedding size. Therefore, the loss function for the prompt-based classifier of each sentence can be expressed as:
\begin{align}
    \mathcal{L}_{class} = \frac{1}{\left\vert\mathcal{S}_{pos}\right\vert}\sum_{i=1}^{\left\vert\mathcal{S}_{pos}\right\vert}-\log\big(p(y|s_i, e_i)\big), \nonumber
\end{align}

% Training and Fine-tuning
\subsection{Training and Fine-tuning}
During the training stage, we sample an episode data from $\mathcal{D}_{train}$ which consists of a support set $\mathcal{\hat{S}}_{train}$ and a query set $\mathcal{\hat{Q}}_{train}$. Unlike previous methods
~\cite{das-etal-2022-container, wang-etal-2022-enhanced,  ma-etal-2022-decomposed, wang-etal-2022-spanproto}, in the training process of PromptNER, we decompose the $\mathcal{\hat{S}}_{train}$ and $\mathcal{\hat{Q}}_{train}$, where the optimized object can be calculated in $\mathcal{\hat{S}}_{train}$ and $\mathcal{\hat{Q}}_{train}$, respectively:
\begin{align}
   % \mathcal{L} = \mathcal{L}_{span} + \mathcal{L}_{class} + \mathcal{L}_{CL}, \nonumber
   \mathcal{L} = \mathcal{L}_{span} + \mathcal{L}_{class}
   \label{ali: train_loss}
\end{align}
During the testing stage, where only label sentences from  $\mathcal{\hat{S}}_{test}$ available, we just use the $\mathcal{\hat{S}}_{test}$ to optimize our model like~(\ref{ali: train_loss}).
% 怎么infer
\subsection{Inference via $k$NN Search}
As described in section~\ref{sec:two_spec}, we denote the collection of candidate spans as $\mathcal{C} = \{(s_i, e_i)_{i = 1}^{3k} \}$. The candidate span embedding is $\mathbf{U}_{query} \in \mathcal{R}^{3k \times d}$, while the prompt label embedding is $\mathbf{U}_{label} \in \mathcal{R}^{t \times d}$. 
Hence, we can compute the probability distribution that the $i$-th span belongs to each class as follows:
\begin{align}
    p_{prompt}(y|s_i, e_i) =  \mathrm{Softmax}(\frac{\mathbf{U}_{label} \mathbf{u}_{i}^\top}{\sqrt{d}}), \nonumber
\end{align}
where $ \mathbf{u}_{i} \in \mathcal{R}^{1 \times d}$, and $d$ is the embedding size. During this inference stage, we filter all the false positive spans which satisfy $\mathrm{none} = \arg\max~p(y|s_i, e_i)$.

To leverage the golden entity representations of the support set $\mathcal{\hat{S}}_{test}$, we introduce the $k$ nearest neighbor search algorithm during the inference stage. First, we merge all the golden entity embedding into a matrix $\mathbf{U}_{golden} \in \mathcal{R}^{n \times d}$, and $n$ is the golden number in the support set $\mathcal{\hat{S}}_{test}$ and $d$ is the embedding size. The similarity score between a candidate span and golden entities is:
\begin{align}
    \mathbf{d}_i = \frac{\mathbf{U}_{golden} \mathbf{u}_{i}^\top}{\sqrt{d}}, \nonumber
\end{align}
where $\mathbf{d}_i \in \mathcal{R}^{n \times 1}$, and $d$ is the embedding size.~Inspired by~\citet{wang2022k}, we just retrieve a golden entity set $\mathcal{N}_i$ with top-$k$ similarity scores. 

\begin{align}
    p(y_i = t|s_i, e_i) \propto \sum_{j=1}^n \mathbb{I}(j \in \mathcal{N}_i, y_j = t) \cdot \mathbf{d}_i(j), \nonumber
\end{align}
where $\mathbb{I}$ is the indicator function. The probability of the label not being retrieved by the $k$-NN search always is assigned as zero.

The final prediction probability is calculated as follows:
\begin{align}
    p(y|s_i, e_i) &= \gamma \cdot  \mathrm{Sigmoid}(\mathbf{R}(s_i, e_i))   \nonumber
                 \\& + \alpha \cdot p_{prompt}(y|s_i, e_i)     \nonumber
                 \\& + \beta \cdot p_{knn}(y|s_i, e_i)
         \label{rerank}
\end{align}
where $\gamma, \alpha, \beta$ are hyper-parameters which balance these three different distributions. The reason why we use $\mathbf{R}(s_i, e_i)$ to rerank is to filter some false positive spans extracted from the position-aware biaffine module.

The final prediction label of span($s_i, e_i$) is:
\begin{align}
    y_{pred} = \mathrm{argmax}~p(y=t|s_i, e_i), \nonumber
\end{align}

\section{Experiment Setup}
\subsection{Datasets}
To demonstrate the few-shot learning ability of our method, we conduct experiments on two well-designed $N$-way $K$-shot few-shot NER datasets.
\textbf{Few-NERD}~\citet{ding-etal-2021-nerd} propose a human-annotated few-shot NER dataset with 8 coarse-grained and 66 fine-grained entity types from Wikipedia. Because the sampling process becomes gradually stricter to satisfy the $K$-shot setting, therefore, each entity type contains $K\thicksim2K$ samples, which alleviates the sampling limitation in Few-NERD. Few-NERD contains two different settings: Intra and Inter. %For Intra, all the entities in the train/dev/test set belong to different coarse-grained types.
\textbf{CrossNER} CrossNER consists of 4 NER datasets from different domains: CoNLL03~\cite{sang2003introduction}(News), WNUT-2017~\cite{derczynski-etal-2017-results}(Social), GUM~\cite{zeldes2017gum}(Wiki) and OntoNotes~\cite{pradhan-etal-2013-towards}(Mixed). For a fair comparison, we use the sampled $N$-way $K$-shot dataset from~\citet{hou-etal-2020-shot}.

\begin{table*}[t]
    \centering
    \setlength{\tabcolsep}{1mm}
    \resizebox{2\columnwidth}{!}{
    \begin{tabular}{lcccccccccc}
    \toprule
        \multirow{3}{*}{\textbf{Models}} & \multicolumn{5}{c}{\textbf{Intra}} & \multicolumn{5}{c}{\textbf{Inter}}\\
        \cmidrule(lr){2-6} \cmidrule(lr){7-11}
        & \multicolumn{2}{c}{\textbf{1$\sim$2-shot}} & \multicolumn{2}{c}{\textbf{5$\sim$10-shot}} & \multirow{2}{*}{\textbf{Avg.}} & \multicolumn{2}{c}{\textbf{1$\sim$2-shot}} & \multicolumn{2}{c}{\textbf{5$\sim$10-shot}} & \multirow{2}{*}{\textbf{Avg.}}\\
        \cmidrule(lr){2-3}\cmidrule(lr){4-5}  \cmidrule(lr){7-8} \cmidrule(lr){9-10} 
         & 5 way & 10 way & 5 way & 10 way & & 5 way & 10 way & 5 way & 10 way &\\
         \cmidrule(lr){1-1}\cmidrule(lr){2-6} \cmidrule(lr){7-11}
         ProtoBERT$^{\dag}$ & 20.76\small\small{\textpm0.84} & 15.05\small{\textpm0.44} & 42.54\small{\textpm0.94} & 35.40\small{\textpm0.13} & 28.44 & 38.83\small{\textpm1.49} & 32.45\small{\textpm0.79} & 58.79\small{\textpm0.44} & 52.92\small{\textpm0.37} & 45.75\\
         NNShot$^{\dag}$ & 25.78\small{\textpm0.91} & 18.27\small{\textpm0.41} & 36.18\small{\textpm0.79} & 27.67\small{\textpm1.06} & 26.98 & 54.29\small{\textpm0.40} & 46.98\small{\textpm1.96} & 50.56\small{\textpm3.33} & 50.00\small{\textpm0.36} & 50.46 \\
         StructShot$^{\dag}$ & 30.21\small{\textpm0.90} & 21.03\small{\textpm1.13} & 38.00\small{\textpm1.29} & 26.42\small{\textpm0.60} & 28.92 & 51.88\small{\textpm0.69} & 43.34\small{\textpm0.10} & 57.32\small{\textpm0.63} & 49.57\small{\textpm3.08} & 50.53\\
         CONTAINER$^{\ddag}$  & 40.43 & 33.84 & 53.70 & 47.49 & 43.87 & 55.95 & 48.35 & 61.83 & 57.12 & 55.81\\
         ESD & 36.08\small{\textpm1.60} & 30.00\small{\textpm0.70} & 52.14\small{\textpm1.50} & 42.15\small{\textpm2.60} & 40.09 & 59.29 \small{\textpm1.25} & 52.16\small{\textpm0.79} & 69.06\small{\textpm0.80} & 64.00\small{\textpm0.43} & 61.13\\
         DecomposedMetaNER & 49.48\small\small{\textpm0.85} & 42.84\small{\textpm0.46} & 62.92\small{\textpm0.57} & 53.14\small{\textpm0.25} & 52.10 &64.75\small{\textpm0.35} & 58.65\small{\textpm0.43} & 71.49\small{\textpm0.47} & 68.11\small{\textpm0.05} & 65.75 \\
         \rowcolor{gray!20}
         \textbf{Ours} & \textbf{55.32{\textpm1.03}} & \textbf{50.29{\textpm0.61}} & \textbf{67.26\small{\textpm1.02}} & \textbf{60.42\small{\textpm0.73}} & \textbf{58.32} & \textbf{64.92{\textpm0.71}} & \textbf{62.28\textpm{0.39}} & \textbf{72.64\small{\textpm0.16}} & \textbf{70.13\small{\textpm0.67}} & \textbf{67.49} \\
        \bottomrule
    \end{tabular}
    }
    \caption{F1 scores with standard deviations on Few-NERD for both Inter and Intra settings. $^{\dag}$ denotes the results reported in \citet{ding-etal-2021-nerd} Arxiv V6 Version.  $^{\ddag}$ is the result without standard deviations from~\cite{das-etal-2022-container}. The best results are in \textbf{bold}.}
    \label{tab:performance_comparison_fewnerd}
\end{table*}

\begin{table*}[t]
    \centering
    \setlength{\tabcolsep}{1mm}
    \resizebox{2\columnwidth}{!}{
    \begin{tabular}{lcccccccccc}
    \toprule
        \multirow{2}{*}{\textbf{Models}} & \multicolumn{5}{c}{\textbf{1-shot}} & \multicolumn{5}{c}{\textbf{5-shot}}\\
        \cmidrule(lr){2-6} \cmidrule(lr){7-11}
         & CoNLL03 & GUM & WNUT & OntoNotes & \textbf{Avg.} & CoNLL03 & GUM & WNUT & OntoNotes & \textbf{Avg.}\\
         \cmidrule(lr){1-1}\cmidrule(lr){2-6} \cmidrule(lr){7-11}
         TransferBERT$^{\dag}$ & 4.75\small{\textpm1.42} & 0.57\small{\textpm0.32} & 2.71\small{\textpm0.72} & 3.46\small{\textpm0.54}  & 2.87 & 15.36\small{\textpm2.81} & 3.62\small{\textpm0.57} & 11.08\small{\textpm0.57} & 35.49\small{\textpm7.60} & 16.39 \\
         SimBERT$^{\dag}$ & 19.22\small{\textpm0.00} & 6.91\small{\textpm0.00} & 5.18\small{\textpm0.00} & 13.99\small{\textpm0.00}  & 11.33 & 32.01\small{\textpm0.00} & 10.63\small{\textpm0.00} & 8.20\small{\textpm0.00} & 21.14\small{\textpm0.00} & 18.00 \\
         Matching Network$^{\dag}$ & 19.50\small{\textpm0.35} & 4.73\small{\textpm0.16} & 17.23\small{\textpm2.75} & 15.06\small{\textpm1.61} & 14.13 & 19.85\small{\textpm0.74} & 5.58\small{\textpm0.23} & 6.61\small{\textpm1.75} & 8.08\small{\textpm0.47}  & 10.03\\
         ProtoBERT$^{\dag}$ & 32.49\small{\textpm2.01} & 3.89\small{\textpm0.24} & 10.68\small{\textpm1.40} & 6.67\small{\textpm0.46}  & 13.43 & 50.06\small{\textpm1.57} & 9.54\small{\textpm0.44} & 17.26\small{\textpm2.65} & 13.59\small{\textpm1.61}  & 22.61\\
         L-TapNet+CDT$^{\dag}$ & 44.30\small{\textpm3.15} & 12.04\small{\textpm0.65} & 20.80\small{\textpm1.06} & 15.17\small{\textpm1.25}  & 23.08 & 45.35\small{\textpm2.67} & 11.65\small{\textpm2.34} & 23.30\small{\textpm2.80} & 20.95\small{\textpm2.81}  & 25.32\\
         DecomposedMetaNER & 46.09\small{\textpm0.44} & 17.54\small{\textpm0.98} & 25.14\small{\textpm0.24} & 34.13\small{\textpm0.92} & 30.73 & 58.18\small{\textpm0.87} & 31.36\small{\textpm0.91} & \textbf{31.02\small{\textpm1.28}} & 45.55\small{\textpm0.90} & 41.53\\
         \rowcolor{gray!20}
         \textbf{Ours} & \textbf{49.69\small{\textpm2.70}} & \textbf{26.24\small{\textpm1.21}} & \textbf{28.07\small{\textpm0.48}} & \textbf{35.38\small{\textpm0.58}} & \textbf{34.85} & \textbf{63.47\small{\textpm1.28}} & \textbf{44.54\small{\textpm0.29}} & 30.40\small{\textpm0.83} & \textbf{48.71\small{\textpm0.59}} & \textbf{46.78} \\
        \bottomrule
    \end{tabular}
    }
    \caption{F1 scores with standard deviations on CrossNER. $^{\dag}$ are the results reported in~\citet{hou-etal-2020-shot}. The best results are in \textbf{bold}.}
    \label{tab:performance_comparison_crossner}
\end{table*}

\subsection{Baselines}
For Few-NERD
~\footnote{
The dataset we used is the newest Few-NERD Arxiv V6 Version. The results of different baselines are shown in \href{https://github.com/microsoft/vert-papers/tree/master/papers/DecomposedMetaNER}{https://github.com/microsoft/vert-papers/tree/master/papers/DecomposedMetaNER}}, we compare PromptNER to CONTaiNER~\cite{das-etal-2022-container}, ESD~\cite{wang-etal-2022-enhanced}, DecomposedNER~\cite{ma-etal-2022-decomposed} and methods from~\citet{ding-etal-2021-nerd}, $e.g.$, StructShot, ProtoBERT, $etc$. For CrossNER, we compare our method to DecomposedNER~\cite{ma-etal-2022-decomposed}, L-TapNet+CDT~\cite{hou-etal-2020-shot} and other methods from~\citet{hou-etal-2020-shot}. We report the micro-F1 scores with standard deviations of different baselines.

\subsection{Implementation Details}
We implement our method using PyTorch version 1.12.1\footnote{\href{https://pytorch.org}{https://pytorch.org}}. We use two separate BERT models for the position-aware biaffine module and the prompt-based classifier, respectively. We load the BERT-base-uncased~\cite{devlin-etal-2019-bert} checkpoint from HuggingFace~\footnote{\href{https://huggingface.co/docs/transformers}{https://huggingface.co/docs/transformers}}. During training, we use the AdamW optimizer with 10\% linear warmup scheduler, and the weight decay ratio is 1e-2. We train our model in the training set and use the validation set to select the model with the highest F1 scores. We also use the AdamW for fine-tuning on the target domain and stop the fine-tuning process early when the loss is less than 1e-2. For more implementation details, please refer to Appendix~\ref{sec:appendix_1}.

\section{Results and Analysis}
\subsection{Main Results}
Table~\ref{tab:performance_comparison_fewnerd} and Table~\ref{tab:performance_comparison_crossner} report the performance of PromptNER on two few-shot NER datasets.
%~\footnote{Noted, we do not compare with~\citet{wang-etal-2022-spanproto} since an unfixed bug of their code will recall all the ground truth entities, whether or not the span detector recognized the ground truth entities.} 
It can be observed that: 
1) Our proposed PromptNER achieves the best performance on Few-NERD and CrossNER. The overall averaged F1 scores over Few-NERD Intra, and Inter setting are improved by 6.22\% and 1.36\% respectively compared to the previous SOTA model DecomposedMetaNER~\cite{ma-etal-2022-decomposed}. Meanwhile, our model also outperforms previous methods by 4.12\% and 5.07\% on CrossNER 1-shot and 5-shot settings, respectively.
2) It is important that we observe the performance improvement on Few-NERD Intra is more significant than on Few-NERD Inter. This phenomenon is because Few-NERD Inter allows the train/dev/test episode to belong to the same coarse-grained types, whereas the train/dev/test episode in Few-NERD Intra must belong to different coarse-grained types and share little knowledge. Therefore, Few-NERD Intra is a more challenging benchmark. The results from Table~\ref{tab:performance_comparison_fewnerd} demonstrate that PromptNER has an excellent transfer learning ability than previous methods when facing difficult tasks.

% ablation
\subsection{Ablation Study}
In this section, we demonstrate the contributions of different parts of Prompt NER. We introduce the following variants for the ablation:
1) Ours w/o Fine-tune
2) Ours w/o Rerank 
3) Ours w/o $k$-NN search
4) Ours w/o Fine-tune and $k$-NN search
5) Ours w/o Position-aware Biaffine
6) Ours w/o Fine-tune and RoPE.
\begin{table}[htb]
\centering
\resizebox{0.9\columnwidth}{!}{
\begin{small}
\begin{tabular}{lcc}
\toprule
\bf Models & \bf Intra & \bf Inter\\
\midrule
\textbf{Ours} & \bf 67.26 & \bf 72.64\\
\cmidrule(lr){1-3}
w/o Fine-tune & 50.99 & 66.64\\
w/o Rerank & 61.79 & 68.18\\
\cmidrule(lr){1-3}
w/o $k$-NN Search  & 65.77 & 71.88 \\
w/o Fine-tune and $k$-NN Search  & 52.45 & 64.41\\
\cmidrule(lr){1-3}
w/o Position-aware Biaffine & 14.23 & 16.43\\
w/o Fine-tune and RoPE & 50.05 & 65.95\\
% \midrule
\bottomrule
\end{tabular}
\end{small}
 }
\caption{Ablation study of different components of our method. We conduct ablation experiments on Few-NERD Intra/Inter 5way 5$\sim$10-shot setting.}
\label{tab:Ablation study}
\end{table}

Results from Tabel~\ref{tab:Ablation study} show that fine-tuning on the novel support set significantly improves the performance of our method. Although we do not fine-tune our model on the novel support set, our method still outperforms all the token-level models in the Few-NERD inter 5way 5$\sim$10 setting, i.e., CONTaiNER~\cite{das-etal-2022-container}, which demonstrates the superiority of our span-based method.
The rerank strategy could also significantly improve the F1 scores of our method, which indicates that this strategy could help to filter some false positive spans.
The $k$-NN Search achieves less performance improvement compared to the model without fine-tuning since fine-tuning the prompt-based classifier on the support set will narrow the embedding distributions between the label word and golden entities in the support set.
According to Table~\ref{tab:Ablation study}, when we remove the Position-aware Biaffine Module during the inference stage, the prompt-based classifier fails to filter the false positives spans, which demonstrates the importance of the span extractor for span-based NER. 
Meanwhile, inserting the absolute and relative position information, i.e., RoPE, into the biaffine module enhances the performance of our method. 
Obviously, the rerank strategy and fine-tuning stage are the key components of our method during inference.
We investigate how the effectiveness of these two components as follows:\\
\textbf{The Effectiveness of Rerank Strategy.}
~The rerank strategy is a crucial component of our method since it could effectively filter some false positive spans. As described in Eq.(\ref{rerank}), we use the scores from the span detector to rerank the final prediction probability. We further investigate how the rerank strategy influences performance. According to Table~\ref{tab:Rerank}, the performance of our method is improved by 5.47\% and 4.46\%, respectively, when applying the rerank strategy during inference. It is worth noting that, although we extract all the spans from the input sentence, the rerank strategy could also significantly improve F1 scores by 38.95\% and 50.23\%, respectively.~This phenomenon indicates that the entities belonging to the category mentioned in the prompt have significantly higher $\mathbf{R}(s_i, e_i)$ scores than entities belonging to other categories, which proves the rerank strategy has the ability to filter some false positive spans.\\
\begin{table}
\centering
\resizebox{1.0\columnwidth}{!}{
\begin{small}
\begin{tabular}{lcc}
\toprule
\bf Models & \bf Intra & \bf Inter\\
\midrule
\textbf{Ours} & \bf 67.26 & \bf 72.64\\
\cmidrule(lr){1-3}
w/o Rerank & 61.79 & 68.18\\
w/o Position-aware Biaffine and Rerank & 14.23 & 16.43\\
w/o Position-aware Biaffine but Rerank & 53.18 & 66.66\\
\bottomrule
\end{tabular}
\end{small}
 }
\caption{The effectiveness of the rerank strategy . We conduct experiments on Few-NERD Intra/Inter 5way 5$\sim$10-shot setting.}
\label{tab:Rerank}
\end{table}

\begin{figure}[!ht]
  \centering
  \includegraphics[width=0.95\columnwidth]{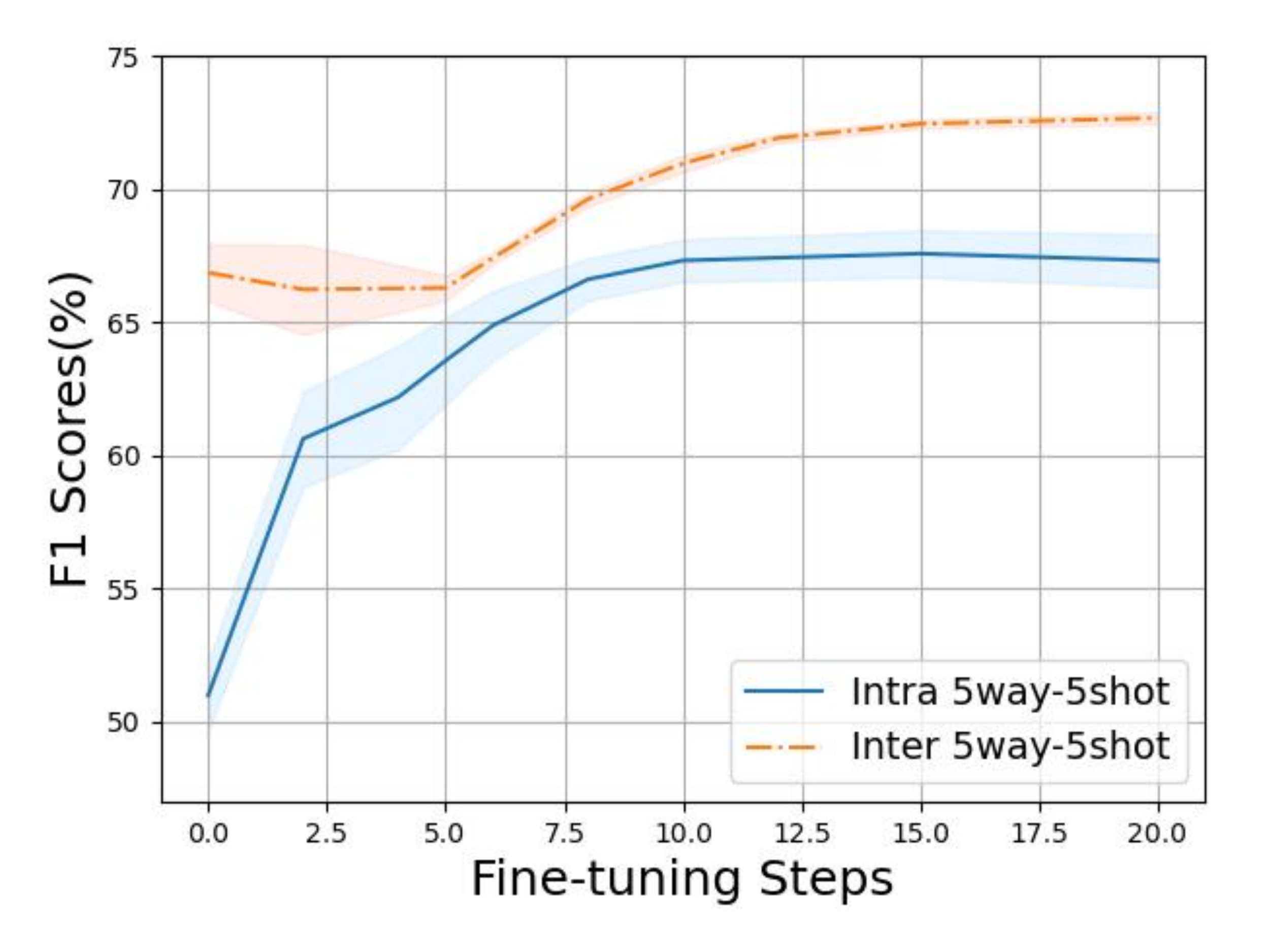} 
  \vspace{-0.3em}
  \caption{The Effectiveness of Fine-tuning. } \label{fig:finetune}
\end{figure}

\textbf{The Effectiveness of Fine-tuning.}
~According to Table~\ref{tab:Ablation study}, fine-tuning on the novel support set during the inference stage will improve the performance by a large margin since our method has no gap between training and fine-tuning. We also investigate how performances are influenced by the different fine-tuning steps. 
As shown in Figure~\ref{fig:finetune}, the performance of our model gradually stabilizes and reaches its peak F1 scores as the number of fine-tuning steps increases, which indicates that our method could effectively utilize the examples from the novel support set to optimize label prototype embeddings.\\

\begin{table}
\centering
\begin{small}
\begin{tabular}{llccc}
\toprule
\bf Settings & \bf N-way & \bf F1 & \bf FP-Span & \bf FP-Type\\
\midrule
\multirow{2}{*}{\textbf{Inter}} & 5-way & 64.92  & 89.24 & 10.76 \\
& 10-way & 62.28  & 81.51 & 18.49 \\
\cmidrule(lr){1-5}
\multirow{2}{*}{\textbf{Intra}} & 5-way & 55.32 & 71.66 & 28.34 \\
& 10-way & 50.29 & 62.31 & 37.69  \\
\bottomrule
\end{tabular}
\end{small}
% }
\caption{Error analysis (\%) of 1$\sim$2-shot settings on Few-NERD dataset. ``FP-Span'' denotes that the span-detector extracts false-positive spans. ``FP-Type'' denotes extracted spans with incorrect entity classes.}
\label{tab:error_analysis}
\end{table}

\begin{table}
\centering
% \resizebox{\linewidth}{!}{
\begin{small}
\begin{tabular}{lccc}
\toprule
\bf Models & \bf F1 & \bf FP-Span & \bf FP-Type  \\
\midrule
ProtoBERT & 38.83 & 86.70 & 13.30 \\
NNShot & 47.24 & 84.70 & 15.30 \\
StructShot & 51.88 & 80.00 & 20.00 \\
ESD & 59.29 & 72.80 & 27.20 \\
DecomMeta & 64.75 & 76.48 & 46.53 \\
\rowcolor{gray!20}
\textbf{Ours} &  \bf64.92  & \bf 89.24 & \bf10.76 \\
% \midrule
\bottomrule
\end{tabular}
\end{small}
% }
\caption{Error analysis (\%) of 5-way 1$\sim$2-shot on Few-NERD Inter for different methods.}
\label{tab:error_analysis_appendix}
\end{table}

\subsection{Error Analysis}
The NER task has two types of errors: `FP-Span' and `FP-Type'.
For FP-Span, it denotes that the span-detector extracts some false-positive spans from the input sentence. 
And FP-Type denotes that the NER system recognizes some true-positive spans but fails to categorize them into correct entity classes. As Table~\ref{tab:error_analysis} shows,  although we conduct the rerank strategy and introduce $\mathrm{none}$ type to filter some false positive spans, PromptNER still tends to extract a few spans with incorrect boundaries. The results from Table~\ref{tab:error_analysis_appendix} also prove that recognizing unseen new class spans only during the span detection stage is difficult for previous few-shot NER systems because current few-shot NER systems are all fully trained in a training set, which results in the few-shot NER system extracting some entities appearing in the training set.
Notably, our method does not follow the traditional prototype networks to use entities representations from the novel support set to construct label prototypes for the span classifier but achieves the lowest FP-Type ratio, demonstrating the superiority of the prompt-based classifier and $k$-NN search over previous traditional prototypical networks for few-shot NER. 

\section{Related Work} 
\subsection{Few-shot Learning and Meta Learning}
Few-shot learning is an essential task that involves learning a model with only a few human-annotated examples~\cite{wang2020generalizing}. In recent years, several methods have been proposed to address different few-shot learning tasks~\cite{geng-etal-2020-dynamic, sheng-etal-2020-adaptive, brown2020language, schick2021s, gao-etal-2021-making} in the NLP community. Meanwhile, various meta-learning algorithms are also proposed to address few-shot learning, i.e., metric learning-based methods~\cite{vinyals2016matching, snell2017prototypical}, optimization-based methods~\cite{finn2017model}, and augmentation-based learning~\cite{ding-etal-2020-daga}.

\subsection{Span-based NER}
Inspired by dependency parsing~\cite{dozat2017deep},~\citet{yu2020named} propose a span-based NER system with a biaffine model. The biaffine model scores each pair of start and end tokens to extract all the candidate spans.To enhance the performance of span-based NER,~\citet{yan2022embarrassingly} use the Convolutional Neural Network (CNN) to utilize spatial relations in the score matrix.~\citet{li-etal-2020-unified} considers the NER task a Machine Reading Comprehension task. Notably, the span-based NER system could handle both flat and nested NER simultaneously, which avoid token-level label dependency problem (i.e, ``BIOES'' rules).  

\subsection{Few-shot NER}
Recently, few-shot NER has received lots of attention in the field of Information Extraction, owing to the high cost of human annotation and the demand for domain-specific knowledge. 
To evaluate the performance of few-shot NER systems better,~\citet{hou-etal-2020-shot} and~\citet{ding-etal-2021-nerd} release two well-designed datasets (CrossNER, Few-NERD) which satisfy the N$\sim$way K$\sim$shot paradigm. 
Research on few-shot NER could be categorized into two types, i.e., one-stage models~\cite{fritzler2019few, hou-etal-2020-shot, tong-etal-2021-learning, das-etal-2022-container} with token-level metric learning, and two-stage models~\cite{yu-etal-2021-shot,wang-etal-2022-enhanced, ma-etal-2022-decomposed, wang-etal-2022-spanproto} following the span-based paradigm. 
Recently,~\citet{wang2022k} has observed that $k$ Nearest Neighbor Search could enhance the performance of the NER system in the low resource scenario.~\citet{ma-etal-2022-label} propose to use pre-trained models to encode the label word to model the label semantics for few-shot NER.~\citet{ming2022few} investigate a novel few-shot nested NER task and design a span-based method to address this problem. 
It is worth noting that all the recent state-of-the-art few-shot NER methods are based on prototypical networks. Previous methods~\cite{das-etal-2022-container, ma-etal-2022-decomposed, wang-etal-2022-spanproto, ming2022few} utilize the support set to construct class prototype representations and use the query set to compute span-level similarities and optimize these label prototype representations. 
However, previous prototype networks are usually unsuitable for fine-tuning in the target domain, where only the support set is available. Different from previous methods, the novelty and contribution of our work are: 
1) We use a prompt to inform PLMs to extract entities and design a prompt-based classifier to conduct span-based metric learning in few-shot NER. 
2) Our method does not use support examples to construct the class prototypes. We use examples from the support set to optimize label prototype embeddings without any gap between the training and fine-tuning stage.
3) Moreover, we introduce the $k$-NN search to enhance the performance of our model. 
4) Our work could also be considered a simple but effective baseline for few-shot NER.

\section{Conclusion}
In this paper, we propose PromptNER, a prompting method for few-shot named entity recognition via $k$ nearest neighbor search. Our approach uses a prompt to instruct Pre-trained Language Models to extract entities with specific classes. We also design a two-stage model with a position-aware biaffine module and a prompt-based classifier with $k$-NN search.
Unlike traditional prototypical networks, our method could use only the novel support set to optimize label prototypes. Extensive experiments demonstrate that our method outperforms previous state-of-the-art few-shot NER methods. Our work provides a novel, simple, and effective baseline for few-shot learning in NER.

\section*{Limitations}
Our proposed method must be trained in a training set for warmup, then utilize its transfer learning ability to address the few-shot NER task. Meanwhile, we also only conduct experiments on the N-way K-shot settings and few-shot flat NER tasks. In the future, we will extend our method to other NER scenarios, such as few-shot nested NER tasks, few-shot Chinese NER tasks.

% \section*{Ethics Statement}

% \section*{Acknowledgements}

\bibliography{anthology,PromptNER}
\bibliographystyle{acl_natbib}

\appendix
\appendix
\clearpage
\section{Appendix} \label{sec:appendix}

\subsection{Implementation Details} \label{sec:appendix_1}
For a fair comparison, we use BERT-base-uncased as the encoder for our method. We use AdamW to optimize our model with 10\% linear warm-up steps. The learning rate of the encoder is 2e-5, and the learning rate of the biaffine decoder is 2e-3. We set the batch size as 1 to narrow the gap between training and fine-tuning, which means we use one episode per step to update our model. For fine-tuning, we stop the fine-tuning process early when the loss is less than 1e-2 or the fine-tuning steps are more than 50. We conduct experiments on Few-NERD and CrossNER with five different random seeds~$\{1~2~3~4~5\}$ and report the average micro-F1 with standard deviations. For inference,  $\gamma, \alpha, \beta$ are hyper-parameters that balance these three distributions. We set $\gamma$ as 0.5 for Few-NERD inter setting and 0.7 for other settings. Meanwhile, we set $\alpha$ as $0.35 * (1 - \gamma)$ and $\beta$ as $0.65 * (1 - \gamma)$, respectively. 
Our source codes are available at~\href{https://github.com/Zhang-Mozhi/PromptNER}{https://github.com/Zhang-Mozhi/PromptNER}.

\subsection{Contrastive Learning} \label{sec:appendix_2}
Recently, Contrastive Learning has been proven effective for token-level metric learning~\cite{das-etal-2022-container}. We also design a span-based contrastive learning algorithm to investigate whether contrastive learning could optimize the span embedding between entities with different labels. In the 1-shot setting, we just let the $X_p$ go through the encoder twice~\cite{gao2021simcse} to obtain sufficient positive samples. We could get the golden span set $\mathcal{M}$ within a support set. Given a golden span $\mathbf{u}_i$, we can define its corresponding positive sample set $\mathcal{M}_i^+$ and in-batch sample set $\mathcal{M}_i^-$:
\begin{align}
    \mathcal{M}_i^+ &= \{\mathbf{u}_j \in \mathcal{M} |y_j = y_i, \mathbf{u}_j \neq \mathbf{u}_i\} ,\nonumber \\
    \mathcal{M}_i^- &= \{\mathbf{u}_j \in \mathcal{M} |y_j \neq y_i, \mathbf{u}_j \neq \mathbf{u}_i\}, \nonumber
\end{align}
Then, the span-based contrastive learning loss can be calculated as follows:
\begin{align}
    \mathcal{L}_{CL} &= -\sum_{i=1}^{\left\vert\mathcal{M}\right\vert}\log \frac{\sum_{(\mathbf{u}_i,  \mathbf{u}_j) \in \mathcal{M}_i^+ }\exp(d(\mathbf{u}_i, \mathbf{u}_j))}{\sum_{ \mathbf{u}_k \in \mathcal{M}_i^- }\exp(\mathrm{d}(\mathbf{u}_i, \mathbf{u}_k))}, \nonumber
\end{align}
where $d$ is a scaled dot product function. By optimizing $\mathcal{L}_{CL}$, we can narrow the embedding distribution of entities with identical labels and separate the entity distribution with different labels. Therefore, the optimized object of PromptNER could be calculated as follows:
\begin{align}
    \mathcal{L} = \mathcal{L}_{span} + \mathcal{L}_{class} + \mathcal{L}_{CL}, \nonumber
\end{align}
Table~\ref{tab:performance_comparison_fewnerd_cl} and Table~\ref{tab:performance_comparison_crossner_cl} denote the performance when applying contrastive learning to our method. We find that contrastive learning will accelerate the overfitting phenomenon of the novel support set, which might harm the performance of our method.

\begin{table*}[ht]
    \centering
    \setlength{\tabcolsep}{1mm}
    \resizebox{2\columnwidth}{!}{
    \begin{tabular}{lcccccccccc}
    \toprule
        \multirow{3}{*}{\textbf{Models}} & \multicolumn{5}{c}{\textbf{Intra}} & \multicolumn{5}{c}{\textbf{Inter}}\\
        \cmidrule(lr){2-6} \cmidrule(lr){7-11}
        & \multicolumn{2}{c}{\textbf{1$\sim$2-shot}} & \multicolumn{2}{c}{\textbf{5$\sim$10-shot}} & \multirow{2}{*}{\textbf{Avg.}} & \multicolumn{2}{c}{\textbf{1$\sim$2-shot}} & \multicolumn{2}{c}{\textbf{5$\sim$10-shot}} & \multirow{2}{*}{\textbf{Avg.}}\\
        \cmidrule(lr){2-3}\cmidrule(lr){4-5}  \cmidrule(lr){7-8} \cmidrule(lr){9-10} 
         & 5 way & 10 way & 5 way & 10 way & & 5 way & 10 way & 5 way & 10 way &\\
         \cmidrule(lr){1-1}\cmidrule(lr){2-6} \cmidrule(lr){7-11}
         ProtoBERT$^{\dag}$ & 20.76\small\small{\textpm0.84} & 15.05\small{\textpm0.44} & 42.54\small{\textpm0.94} & 35.40\small{\textpm0.13} & 28.44 & 38.83\small{\textpm1.49} & 32.45\small{\textpm0.79} & 58.79\small{\textpm0.44} & 52.92\small{\textpm0.37} & 45.75\\
         NNShot$^{\dag}$ & 25.78\small{\textpm0.91} & 18.27\small{\textpm0.41} & 36.18\small{\textpm0.79} & 27.67\small{\textpm1.06} & 26.98 & 54.29\small{\textpm0.40} & 46.98\small{\textpm1.96} & 50.56\small{\textpm3.33} & 50.00\small{\textpm0.36} & 50.46 \\
         StructShot$^{\dag}$ & 30.21\small{\textpm0.90} & 21.03\small{\textpm1.13} & 38.00\small{\textpm1.29} & 26.42\small{\textpm0.60} & 28.92 & 51.88\small{\textpm0.69} & 43.34\small{\textpm0.10} & 57.32\small{\textpm0.63} & 49.57\small{\textpm3.08} & 50.53\\
         CONTAINER$^{\ddag}$  & 40.43 & 33.84 & 53.70 & 47.49 & 43.87 & 55.95 & 48.35 & 61.83 & 57.12 & 55.81\\
         ESD & 36.08\small{\textpm1.60} & 30.00\small{\textpm0.70} & 52.14\small{\textpm1.50} & 42.15\small{\textpm2.60} & 40.09 & 59.29 \small{\textpm1.25} & 52.16\small{\textpm0.79} & 69.06\small{\textpm0.80} & 64.00\small{\textpm0.43} & 61.13\\
         DecomposedMetaNER & 49.48\small\small{\textpm0.85} & 42.84\small{\textpm0.46} & 62.92\small{\textpm0.57} & 53.14\small{\textpm0.25} & 52.10 &64.75\small{\textpm0.35} & 58.65\small{\textpm0.43} & 71.49\small{\textpm0.47} & 68.11\small{\textpm0.05} & 65.75 \\
         \rowcolor{gray!20}
         \textbf{Ours w/o CL} & \textbf{55.32{\textpm1.03}} & \textbf{50.29{\textpm0.61}} & \textbf{67.26\small{\textpm1.02}} & \textbf{60.42\small{\textpm0.73}} & \textbf{58.32} & 64.92{\textpm0.71} & \textbf{62.28\textpm{0.39}} & \textbf{72.64\small{\textpm0.16}} & \textbf{70.13\small{\textpm0.67}} & \textbf{67.49} \\
         \rowcolor{gray!20}
         \textbf{Ours w CL} & 54.92{\textpm0.56} & 49.49{\textpm0.54}& 66.97\small{\textpm0.10} & 59.77\small{\textpm0.65} & 57.79 & \textbf{64.93{\textpm0.44}} & 62.16\textpm{0.36} & 72.15\small{\textpm0.20} & 69.20\small{\textpm0.89} & 67.11 \\
        \bottomrule
    \end{tabular}
    } 
    \caption{F1 scores with standard deviations on Few-NERD for both Inter and Intra settings. $^{\dag}$ denotes the results reported in \citet{ding-etal-2021-nerd} Arxiv V6 Version.  $^{\ddag}$ is the result without standard deviations from~\cite{das-etal-2022-container}. The best results are in \textbf{bold}.}
    \label{tab:performance_comparison_fewnerd_cl}
\end{table*}

\begin{table*}[ht]
    \centering
    \setlength{\tabcolsep}{1mm}
    \resizebox{2\columnwidth}{!}{
    \begin{tabular}{lcccccccccc}
    \toprule
        \multirow{2}{*}{\textbf{Models}} & \multicolumn{5}{c}{\textbf{1-shot}} & \multicolumn{5}{c}{\textbf{5-shot}}\\
        \cmidrule(lr){2-6} \cmidrule(lr){7-11}
         & CoNLL03 & GUM & WNUT & OntoNotes & \textbf{Avg.} & CoNLL03 & GUM & WNUT & OntoNotes & \textbf{Avg.}\\
         \cmidrule(lr){1-1}\cmidrule(lr){2-6} \cmidrule(lr){7-11}
         TransferBERT$^{\dag}$ & 4.75\small{\textpm1.42} & 0.57\small{\textpm0.32} & 2.71\small{\textpm0.72} & 3.46\small{\textpm0.54}  & 2.87 & 15.36\small{\textpm2.81} & 3.62\small{\textpm0.57} & 11.08\small{\textpm0.57} & 35.49\small{\textpm7.60} & 16.39 \\
         SimBERT$^{\dag}$ & 19.22\small{\textpm0.00} & 6.91\small{\textpm0.00} & 5.18\small{\textpm0.00} & 13.99\small{\textpm0.00}  & 11.33 & 32.01\small{\textpm0.00} & 10.63\small{\textpm0.00} & 8.20\small{\textpm0.00} & 21.14\small{\textpm0.00} & 18.00 \\
         Matching Network$^{\dag}$ & 19.50\small{\textpm0.35} & 4.73\small{\textpm0.16} & 17.23\small{\textpm2.75} & 15.06\small{\textpm1.61} & 14.13 & 19.85\small{\textpm0.74} & 5.58\small{\textpm0.23} & 6.61\small{\textpm1.75} & 8.08\small{\textpm0.47}  & 10.03\\
         ProtoBERT$^{\dag}$ & 32.49\small{\textpm2.01} & 3.89\small{\textpm0.24} & 10.68\small{\textpm1.40} & 6.67\small{\textpm0.46}  & 13.43 & 50.06\small{\textpm1.57} & 9.54\small{\textpm0.44} & 17.26\small{\textpm2.65} & 13.59\small{\textpm1.61}  & 22.61\\
         L-TapNet+CDT$^{\dag}$ & 44.30\small{\textpm3.15} & 12.04\small{\textpm0.65} & 20.80\small{\textpm1.06} & 15.17\small{\textpm1.25}  & 23.08 & 45.35\small{\textpm2.67} & 11.65\small{\textpm2.34} & 23.30\small{\textpm2.80} & 20.95\small{\textpm2.81}  & 25.32\\
         DecomposedMetaNER & 46.09\small{\textpm0.44} & 17.54\small{\textpm0.98} & 25.14\small{\textpm0.24} & 34.13\small{\textpm0.92} & 30.73 & 58.18\small{\textpm0.87} & 31.36\small{\textpm0.91} & \textbf{31.02\small{\textpm1.28}} & 45.55\small{\textpm0.90} & 41.53\\
         \rowcolor{gray!20}
         \textbf{Ours w/o CL} & \textbf{49.69\small{\textpm2.70}} & \textbf{26.24\small{\textpm1.21}} & \textbf{28.07\small{\textpm0.48}} & \textbf{35.38\small{\textpm0.58}} & \textbf{34.85} & \textbf{63.47\small{\textpm1.28}} & \textbf{44.54\small{\textpm0.29}} & 30.40\small{\textpm0.83} & 48.71\small{\textpm0.59} & \textbf{46.78} \\
         \rowcolor{gray!20}
         \textbf{Ours w CL} & 46.37\small{\textpm3.55} & 24.46\small{\textpm1.55} & 27.03\small{\textpm0.98} & 33.48\small{\textpm0.47} & 32.84 & 63.29\small{\textpm1.84} & 43.14\small{\textpm1.09} & 30.17\small{\textpm0.67} & \textbf{48.75\small{\textpm1.12}} & 46.34 \\
        \bottomrule
    \end{tabular}
    }
    \caption{F1 scores with standard deviations on CrossNER. $^{\dag}$ are the results reported in~\citet{hou-etal-2020-shot}. The best results are in \textbf{bold}.}
    \label{tab:performance_comparison_crossner_cl}
\end{table*}

\end{document}